\documentclass[lettersize,journal]{IEEEtran}
\usepackage{amsmath,amsfonts}
\usepackage{algpseudocodex}
\usepackage{algorithm}
\usepackage{array}
\usepackage[caption=false,font=normalsize,labelfont=sf,textfont=sf]{subfig}
\usepackage{textcomp}
\usepackage{stfloats}
\usepackage{url}
\usepackage{verbatim}
\usepackage{graphicx}
\usepackage{cite}
\usepackage[colorlinks,bookmarksnumbered,citecolor=orange,urlcolor=orange]{hyperref}
 \hypersetup{
     colorlinks=true,
     linkcolor=orange,
     filecolor=orange,
     citecolor=orange,      
     urlcolor=orange,
     }
\usepackage{caption}
\usepackage{amssymb}
\usepackage[table]{xcolor}
\usepackage{pifont}
\usepackage{booktabs}
\usepackage{caption}

\usepackage{enumitem}

\newcommand{\p}[1]{\smallskip \noindent \textbf{{#1}.}}
\newcommand{\eq}[1]{Equation~(\ref{eq:#1})}
\newcommand{\fig}[1]{Figure~\ref{fig:#1}}

\usepackage{balance}

\usepackage{longtable}
\usepackage{array,etoolbox}
\preto\tabular{\setcounter{magicrownumbers}{0}}
\newcounter{magicrownumbers}

\usepackage{listings}
\usepackage{mdframed}
\usepackage{xcolor}
\lstnewenvironment{SystemBox}[1]{%
    \subsubsection{#1}
    \mdframed[
        backgroundcolor=gray!10,
        hidealllines=true, 
        innerleftmargin=12pt, 
        innertopmargin=1pt,
        innerbottommargin=1pt
    ]
    \lstset{
        basicstyle=\ttfamily\small,
        breaklines=true,
        numbers=left,
        numberstyle=\tiny\color{gray},
        numbersep=5pt,
        xleftmargin=1pt,
        frame=none,
        columns=fullflexible,
        keepspaces=true,
        escapechar=\#
    }
}{%
    \endmdframed
}

\begin{document}

\definecolor{myblue}{HTML}{4C72B0}
\definecolor{myorange}{HTML}{DD8452}
\definecolor{mygreen}{HTML}{55A868}
\definecolor{myred}{HTML}{C44E52}

\title{Language Movement Primitives: \\Grounding Language Models in Robot Motion}


\author{Yinlong Dai$^1$, Benjamin A. Christie$^1$, Daniel J. Evans$^1$, Dylan P. Losey$^1$, and Simon Stepputtis$^2$
\thanks{This work is supported in part by NSF Grant $\#2337884$.
}
\thanks{$^1$\href{https://collab.me.vt.edu}{Collab}, Dept. of Mechanical Engineering, Virginia Tech, Blacksburg, VA 24061. \texttt{\{daiyinlong, benc00, danielevans, losey\}@vt.edu}}
\thanks{$^2$\href{https://tealab.ai}{TEA Lab}, Dept. of Mechanical Engineering, Virginia Tech, Blacksburg, VA 24061. \texttt{stepputtis@vt.edu}}
}



\maketitle

\begin{abstract}
Enabling robots to perform novel manipulation tasks from natural language instructions remains a fundamental challenge in robotics, despite significant progress in generalized problem solving with foundational models.
Large vision and language models (VLMs) are capable of processing high-dimensional input data for visual scene and language understanding, as well as decomposing tasks into a sequence of logical steps; however, they struggle to ground those steps in embodied robot motion.
On the other hand, robotics foundation models output action commands, but require in-domain fine-tuning or experience before they are able to perform novel tasks successfully.
At its core, there still remains the fundamental challenge of connecting abstract task reasoning with low-level motion control.
To address this disconnect, we propose Language Movement Primitives (LMPs), a framework that grounds VLM reasoning in Dynamic Movement Primitive (DMP) parameterization.
Our key insight is that DMPs provide a small number of interpretable parameters, and VLMs can set these parameters to specify diverse, continuous, and stable trajectories.
Put another way: VLMs can reason over free-form natural language task descriptions, and semantically ground their desired motions into DMPs --- bridging the gap between high-level task reasoning and low-level position and velocity control.
Building on this combination of VLMs and DMPs, we formulate our LMP pipeline for zero-shot robot manipulation that effectively completes tabletop manipulation problems by generating a sequence of DMP motions.
Across 31 real-world manipulation tasks, we show that LMP achieves 65\% task success as compared to 35\% for the best performing baseline. See videos at our website: \url{https://collab.me.vt.edu/lmp}
\end{abstract}

\begin{IEEEkeywords}
Task and Motion Planning, Manipulation, Hybrid Learning, Neuro-Symbolic Method
\end{IEEEkeywords}

\IEEEpeerreviewmaketitle

\section{Introduction} \label{sec:intro}

\begin{figure}[t]
    \centering
    \includegraphics[width=1.0\linewidth]{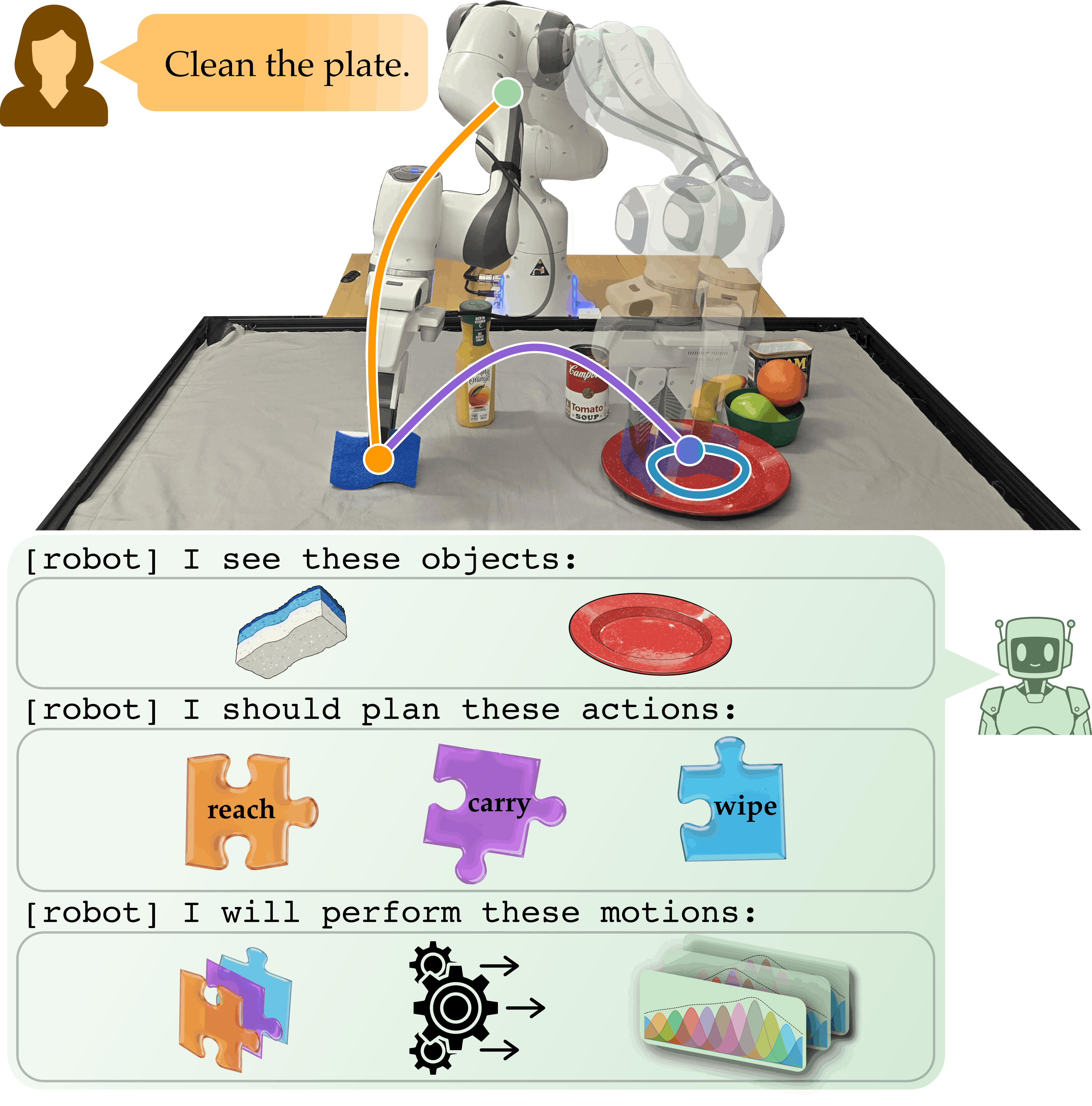}
    \caption{Overview of Language Movement Primitives (LMP). Given a task description from the user, LMP first detects objects in the environment and composes a suitable sequence of high-level subtasks to achieve the overall goal. For each subtask, LMP then generates low-level parameters to define a Dynamic Movement Primitive (DMP). The robot tracks the continuous DMP trajectory, grounding its semantic reasoning for zero-shot robot manipulation.}
    \label{fig:front}
\end{figure}

Robots should be able to perform new tasks given simple user instructions.
For example, imagine that a human tells their robot to ``clean the plate.'' 
Ideally, the robot reasons over this specification and the observed environment, breaks the task down into steps, and then cleans the plate.
What makes this particularly challenging is that the human's command implies motions that are never explained: e.g., picking up a cloth and moving it in a circular pattern.
Consider \fig{front} where the robot realizes that it needs a cleaning implement, grasps a sponge, and then starts wiping.
Unlike prior works in which the robot relies on human demonstrations \cite{chi2025diffusion, dasari2025ingredients, team2024octo} or multiple rounds of real-world experience \cite{kalashnikov2018qtopt, zeng2018learning}, we achieve \textit{zero-shot} generalized intelligence by connecting the human's language specification to the robot's controlled motions.

Our work is related to recent efforts at the intersection of robotics and foundation models and view these efforts along a spectrum.
At one extreme are methods that train models specifically on large-scale \textit{robotic datasets}: these approaches obtain a policy that can directly convert language and vision into actions for the robot to execute \cite{zitkovich2023rt, kim2024openvla, team2024octo, black2024pi_0}.
At the other extreme are methods that train on larger and more diverse \textit{non-robotic datasets} \cite{ye2024latent, li2025scalable} with the goal of creating general-purpose foundation models, allowing these generalized vision and language models to decompose tasks and output actions at a symbolic level (e.g., ``pick up the sponge'').
However, both approaches have critical shortcomings.
Robotics foundation models, while trained on extensive robot data, have limited common-sense reasoning capabilities and still require in-domain fine-tuning~\cite{black2024pi_0} or experience collection~\cite{intelligence2025pi}.
General purpose foundation models are not grounded in robotics, and thus their outputs are not immediately actionable --- while the robot knows \textit{what} to do, it does not know \textit{how} to ``pick up the sponge'' with respect to the robot's joint's actuation.

\IEEEpubidadjcol

Recent neuro-symbolic approaches try to sidestep this gap by separating high-level task reasoning from low-level robot control.
In recent works like \cite{velasquez2025neurosymbolic, saycan2022arxiv, liang2022code, Kwon2024Language} the robot combines a high-level planner --- which provides generalized reasoning --- with a low-level controller --- which converts symbolic plans into robotic motions.
However, these approaches perform symbolic reasoning over \textit{discrete} action primitives rather than the \textit{continuous} motion parameters, constraining the system's ability to perform complex movements.
This means our fundamental bottleneck persists: to ``clean the plate,'' we must establish a robust connection between high-level task reasoning and low-level robotic control for precise position, velocity, and acceleration profiles \cite{khan2025foundation, bai2025embodied}.

In this work we propose to bridge the gap and provide generalized robot policies by connecting vision and language models (VLMs) to Dynamic Movement Primitives (DMPs). We consider DMPs to be a particularly suitable controller formulation for grounding language into robot motion, as they combine semantically meaningful parameters that define the motion, i.e., their forcing function, while providing stable goal-attractor dynamics. 
We specifically focus on table-top manipulation tasks.
Given a user command (e.g., ``clean the plate'', see Fig.~\ref{fig:front}), the VLM outputs task-based reasoning, breaking down the steps and even realizing what is missing from the human's specification.
DMPs offer a control-theoretic way to convert each of these steps into motions, parameterizing trajectories while guaranteeing the robot will reach its goal.
Our insight is that:
\begin{center}
    \textit{DMPs physically ground VLMs because they embed diverse and expressive trajectories into a small number of parameters which the VLM can intuitively tune.}
\end{center}
Our resulting system --- which we refer to as \textbf{Language Movement Primitives (LMPs)} --- integrates off-the-shelf VLMs into a full-stack pipeline for robotic task execution, combining the best of both worlds by leveraging the high-level reasoning capabilities of VLMs with the precise control of DMPs.
During inference, LMP uses the human's task description and an RGB-D image of the robot's environment.
By segmenting the input image, we generate a language description of the environment, including object types and locations.
The VLM then reasons over this verbal state to create a high-level plan, and decides on the next subtask towards completing that plan.
More specifically, the VLM chooses the parameters for the next DMP, and the robot executes this DMP in the environment.
If a failure occurs, we incorporate a refinement loop where the human can provide corrective feedback (e.g., ``use a towel instead''), and the robot iteratively refines its motion controller for subsequent attempts.

Overall, we see this paradigm as a step towards general-purpose robot arms that can be controlled by everyday humans without requiring additional demonstrations. 
We make the following contributions:

\p{Formalizing LMPs}
We frame Language Movement Primitives as an abstracted policy.
The \textit{state} is the image and generated text description of the environment, and the \textit{action} is the weights of the next DMP.
The human's task specification is an implicit reward that the policy can leverage to choose actions, providing a semantically meaningful connection between high-level VLM planning and low-level DMP control.

\p{Grounding Language to Control} 
We present a complete system that translates open-form user instructions and corrective feedback into fine-grained and stable low-level motion controllers. 
Our user interface accepts commands specified in natural language, and also supports effortless and intuitive user feedback without requiring detailed knowledge of robot kinematics or low-level control specifications. 

\p{Comparing to Foundation Models}
We compare LMPs to foundation models, such as $\pi_{0.5}$~\cite{intelligence2025pi_}, in robotics and other neuro-symbolic approaches across $31$ real-world manipulation tasks.
Without any fine-tuning or feedback, LMP outperforms the best baseline by $19\%$ ($54\%$ vs. $35\%$); with up to three rounds of natural-language feedback, the gap grows to $30\%$.
As part of our tests, we evaluate on parts of the CALVIN benchmark in simulation, as well as 31 real-world tabletop manipulation tasks across static and dynamically changing environments.
Performance particularly improves in scenarios that require trajectory shaping for obstacle avoidance or multi-stage tasks.
\section{Related Works}\label{sec:related}

Learning general-purpose robot policies remains an open challenge. 
Recent approaches range from small expert models tailored on specific robots and tasks~\cite{brohan2022rt, lynch2021languageconditionedimitationlearning, stepputtis2020language} to large foundation models for cross-embodiment~\cite{black2024pi_0, Kawaharazuka_2025}.
Complementing these data-driven methods, neuro-symbolic approaches combine LLM reasoning with planning and motion control through symbolic representations~\cite{bhagat2024let, keller2025neurosymbolicimitationlearningdiscovering}.
In the following subsections, we discuss these three approaches in more detail while positioning our proposed method alongside these approaches. 

\subsection{Small Expert Models for Robot Learning}
Control-theoretic approaches such as \cite{pmlr-v78-campbell17a} and \cite{schaal2006dynamic} learn behaviors for particular tasks and robots with only a few examples, but struggle beyond their training distribution.
Deep learning has demonstrated strong results for robot policy and multi-task learning by modeling complex action distributions~\cite{chi2025diffusion, shafiullah2022behavior}, whether through imitation~\cite{xie2020deep, zhang2018deep} or reinforcement learning~\cite{haarnoja2024learning, ibarz2021train}.
While effective, demonstration-centric paradigms impose a substantial burden on the user to provide suitable examples \cite{ravichandar2020recent}, while reinforcement learning methods require large-scale training due to the complexity of real-world environments~\cite{luo2024serl}.
Hybrid approaches that initialize policies with imitation or reinforcement learning and subsequently refine them through experience~\cite{ross2011dagger, elelimy2025rethinking, hu2025continual} partially address this issue; however, generalization to novel scenario remains limited.
World models offer an alternative by enabling robots to plan with scenario roll-outs~\cite{hafner2023mastering, zhou2024dino}, but remain computationally expensive and scale poorly to high-dimensional real-world environments.

A core challenge across these approaches --- particularly in household and tabletop manipulation --- is the high variety of tasks that need to be learned and the associated difficulty of obtaining sufficient training data~\cite{khazatsky2024droid, dai2025prepare}.
Open X-Embodiment~\cite{o2024open} addresses this data scarcity challenge by aggregating datasets from different real-world robot tasks, allowing for positive transfer across embodiments within similar task domains. 
In contrast, LMP leverages the \textit{reasoning capabilities of large foundation models} to generate task-specific robot controllers through a semantically interpretable parameter space, \textit{removing the need for demonstrations}. 

\subsection{Robotics Foundation Models}
Internet-scale foundation models have enabled progress in robot
task planning~\cite{dalal2025local}, multimodal perception~\cite{kim2024openvla}, and zero-shot adaptation to novel tasks and environments~\cite{liang2022code,team2025gemini}. 
Specifically, Vision-Language-Action (VLA) models~\cite{kim2024openvla, black2024pi_0} leverage vision and language pipelines to ground high-dimensional data in a shared embedding space, conditioning downstream action generation~\cite{stepputtis2020language}. 
RT-2~\cite{zitkovich2023rt} popularized this VLA paradigm by co-fine-tuning vision-language models on robotic trajectory data, enabling emergent capabilities such as chain-of-thought reasoning and semantic understanding that transfers from internet-scale pre-training.
Other methods prompt language models to output code that generates trajectories~\cite{liang2022code}, while related works leverage LLMs as evaluators or reward models to provide feedback for policy learning~\cite{rocamonde2023vision, saycan2022arxiv}. 
Despite these advances, a fundamental gap remains in \textit{grounding high-level semantic reasoning into low-level continuous control while generalizing across diverse tasks}.
LMP addresses this gap by providing a semantically meaningful parameter space such that large-scale foundation models can effectively generate low-level controllers while performing common-sense task reasoning across the control parameters.

\subsection{Hybrid Models}
Neuro-symbolic and hybrid systems have shown success by leveraging symbolic reasoning and planning as an integral part of the robot's decision making and motion generation process~\cite{velasquez2025neurosymbolic, dalal2025local, stepputtis2020language}.
Recent approaches incorporate neural vision and language-based components into symbolic pipelines, enabling robots to reason over unstructured sensory inputs while maintaining structured task representations ~\cite{bhagat2024let, liang2024visualpredicator}.
For example, \cite{capitanelli2024framework} employ LLMs as neuro-symbolic task planners compatible with standard planning approaches, applying their generative capabilities to overcome limitations of traditional symbolic planners in dynamic human-robot collaboration scenarios.
Leveraging LLMs enables robots to perform complex task decompositions, semantic grounding, and symbolic planning, while relying on learned controllers for execution of atomic task primitives~\cite{saycan2022arxiv, li2024shapegrasp}. 
However, \textit{existing hybrid approaches typically rely on hand-designed symbolic abstractions or a discrete library of learned controllers that are tightly coupled to specific robots and tasks}. 

Beyond task planning, structured motion representation through control-theoretic approaches, such as DMP, has been used to model robot behavior~\cite{schaal2006dynamic, saveriano2023dynamic}, offering a foundation with convergence guarantees while modeling motion through learned forcing functions. 
Recent extensions have combined DMPs with reinforcement learning~\cite{li2024efficient}, demonstrating their versatility for trajectory representation. 
Our LMP framework is inspired by these directions, leveraging a large language model for high-level reasoning, generating a set of parameters for a low-level DMP motion controller, bridging the gap between language understanding, common-sense reasoning, and fine-grained motion control.


Prior work, such as~\cite{zhou2023generalizable}, uses LLMs to sequence predefined DMPs through symbolic pre/post-conditions, requiring the fitting DMPs to demonstrations of potential motions and object interactions.
However, instead of sequencing existing policies, we propose to leverage the spatial reasoning capabilities of LLM to directly generate DMP weights and parameters, which is a largely unexplored problem.
Our proposed method, LMP, is a demonstration-free zero-shot framework that directly generates DMP parameters from language and visual observations. Through our results, we show that LLMs find DMP weights semantically interpretable, while the expressive structure of DMPs enables these generated parameters to be physically grounded into executable robot motions. 
Furthermore, LMP introduces a unified pipeline of task decomposition, controller generation, and iterative refinement for robot motion generation.
\begin{figure*}[t]
    \centering
    \includegraphics[width=\linewidth]{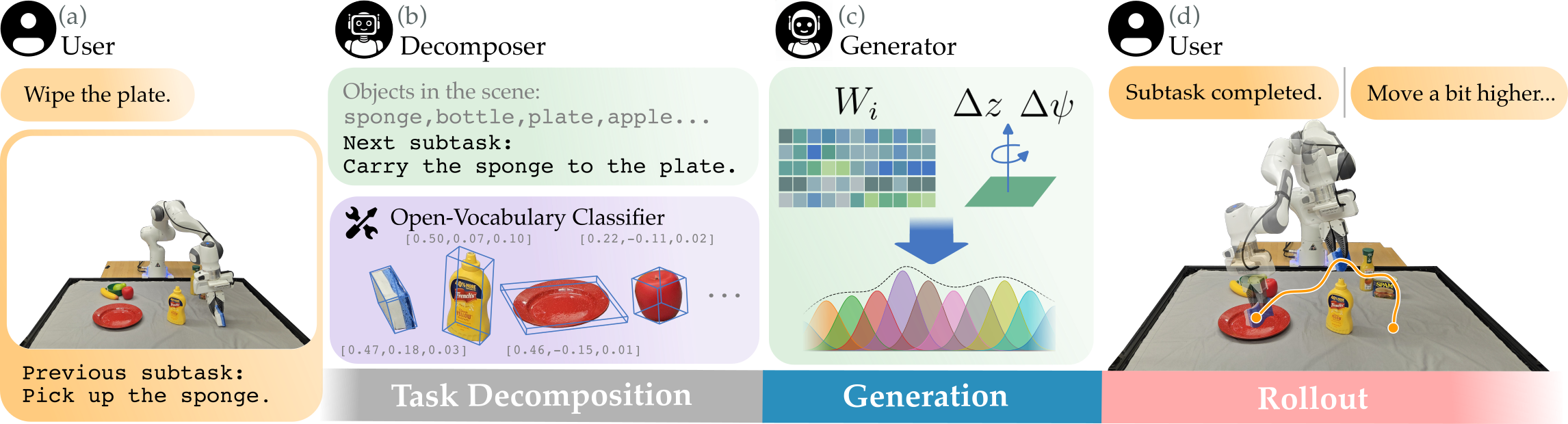}
    \vspace{-1.5em}    
    \caption{LMP pipeline for a single subtask rollout. \textbf{(a)} The robot begins with a user-provided task description. The robot then collects an image capturing the current environment state, and remembers any previously performed subtask(s). \textbf{(b)} The decomposer $\pi_\mathcal{D}$ identifies scene objects and outputs a subtask for the next DMP to complete. An open-vocabulary classifier and depth sensing are used to estimate 3D object locations. The scene description and proposed subtask are then forwarded to the DMP weight generator $\pi_\mathcal{G}$. \textbf{(c)} The generator predicts DMP weights and auxiliary parameters to define the low-level reference trajectory. \textbf{(d)} The robot tracks the continuous trajectory generated from the predicted DMP parameters. Optionally, the user may observe the robot and provide natural-language feedback about any mistakes. If the user gives this refinement $r$, then the robot resets the rollout and the process repeats from (b).}
    \label{fig:method}
\end{figure*}

\section{Problem Statement}\label{sec:problem}

We consider settings where a robot arm is performing tabletop manipulation tasks.
The human operator provides a natural language description of the task $\tau$ they want the robot to perform, and the robot converts this description into task execution (e.g., embodied motion).
The robot does not have access to any in-domain demonstrations.

\p{Observation} 
Let $o \in \mathcal{O}$ be the robot's observation of its environment.
In our experiments $o$ contains the robot's joint position (measured by onboard encoders) and an RGB-D image of the environment (taken using an external depth camera).
This observation captures where the robot is, and also perceives the different objects on the table.

\p{Policy}
The robot converts the user's task description $\tau \in \mathcal{T}$ and the environment observation $o$ into actions:
\begin{equation} \label{eq:policy1}
    a \sim \pi\left(\circ \mid \tau, {o}\right)
\end{equation}
Here $\pi$ is a \textit{high-level} policy: the actions produced by this policy are not low-level position or joint torque commands.
Instead, we treat the actions as motion primitives (with details below), and rely on the high-level policy $\pi$ to select the next motion primitive that will help complete the given task $\tau$.

\p{Action}
We define the robot's actions $a \in \mathcal{A}$ as parameterized motions.
Related neuro-symbolic works have instantiated actions as absolute positions \cite{Kwon2024Language}, learned skills \cite{dalal2025local}, or a library of general-purpose motions \cite{gao2024physically}.
We instead use Dynamic Movement Primitives (DMPs) as the \textit{actions} that connect vision and language models to embodied behaviors.

\p{Dynamic Movement Primitives (DMP)} 
DMPs parameterize a smooth motion between two points.
Let $\xi$ be a single degree of freedom in the robot's end-effector space; e.g., $\xi$ could be the robot's position along the $x$ axis.
DMPs parameterize the trajectory $\xi(t)$ between the start state $\xi(0)$ and the goal state $g$ as a second-order linear dynamical system \cite{schaal2006dynamic, li2023prodmp}:
\begin{equation}
    T^{2}\ddot{\xi} = \alpha\bigl(\beta (g - \xi) - T \dot{\xi}\,\bigr) + f(z)
\end{equation}
Here $\xi$ is position, $\dot{\xi}$ is velocity, and $\ddot{\xi}$ is acceleration.
The temporal scaling parameter $T$ modulates the execution speed of the trajectory, while $\alpha$ and $\beta$ are positive constants that define a spring-damper system (regulating how quickly the trajectory is attracted to the goal position $g$).


We can shape the trajectory of the DMP using the nonlinear forcing function $f(z)$.
In our experiments we instantiate this forcing function as a normalized weighted sum of Gaussian basis functions $\{\psi_i\}_{i=1}^N$ parameterized by the phase variable $\tilde t$ and modulated by the canonical decay variable $z \in [0,1]$:
\begin{equation} \label{eq:P3}
f(z) =
\frac{z(t)\sum_{i=1}^{N} w_i \, \psi_i(\tilde t\,)}
     {\sum_{i=1}^{N} \psi_i(\tilde t\,)}
     ,\quad
    \tilde t = \min\!\left(\frac{t}{T},\,1\right)
\end{equation}
Each Gaussian basis function is given by:
\begin{equation} \label{eq:P4}
    \psi_i(\tilde t) = \exp\!\left(-h (\tilde t - c_i)^2 \right)
\end{equation}
where the basis centers $c_i$ are uniformly distributed over $[0,1]$ and $h = 1/\sigma^2$ controls the basis width. 

Within our work the weights $w$ are particularly important.
Element $w_i$ determines the relative impact of the $i$-th basis function; by increasing or decreasing these weights, we modulate the shape of trajectory $\xi(t)$ along a continuous spectrum.
Indeed, we can specify the shape of the entire motion using weights $w$ and goal position $g$.
When using DMPs, the robot is guaranteed to converge to the goal as the canonical variable $z \rightarrow 0$.
We update $z$ as a time-dependent decay function of the normalized time $\tilde t$:
\begin{equation}
    z(t) = \exp\!\left(-\gamma \tilde t^{\,3}\right)
\end{equation}
where $\gamma > 0$ controls $z$'s rate of decay. 
In practice, $z$ remains close to one until the end of the trajectory: when $z \rightarrow 1$, the forcing function can alter the motion profile.
As $z \rightarrow 0$, the nonlinear terms are ignored and the robot converges to $g$.

Overall, DMPs serve as motion primitives that capture a wide range of trajectories through a small number of weights $w$ and goals $g$. 
We know that trajectory $\xi(t)$ will converge to the goal $g$, and directly adjust the trajectory shape using weights $w$.
DMPs enable rich task-specific motion shaping without sacrificing stability.
Within the context of our high-level policy in \eq{policy1}, the set of DMP weights and goals becomes the robot's action space $\mathcal{A}$.

\p{Subtasks}
At the start of the interaction the human provides a task description $\tau$.
The robot then completes this task in a sequence of steps.
At each step the robot observes $o$, and queries the high-level policy $\pi$ for the weights and goals of the next DMP.
The robot arm rolls out the DMP and interacts with the physical environment.
This process then repeats until the task is complete (outputting a null command).
We refer to the steps as \textit{subtasks} $\tau_i$, and denote the robot's observation and action at each subtask as $o_i$ and $a_i$, respectively.

\p{Feedback}
Ideally, the sequence of actions $[a_i]_{0}^K$ should solve the original task $\tau$.
But we recognize that no policy is perfect and that the generated weights ${w}$ may not complete the task.
To this end, we incorporate additional natural language feedback at the end of each subtask, allowing the robot to re-attempt the task by refining the previously generated weights given verbal feedback.
Alternatively, the human may have \textit{new} instructions they want to provide.
Our problem setting formulates this feedback as a natural language refinements $r$.
We emphasize that this refinement is not necessary, but our framework incorporates it effectively when provided.
\section{Language Movement Primitives (LMPs)} \label{sec:method}

To convert high-level language prompts $\tau$ into low-level control trajectories we propose Language Movement Primitives (LMPs).
The central idea of LMPs is a combination of VLMs and DMPs.
The VLM serves as the high-level policy from \eq{policy1}: this model takes in segmented state observations, breaks the task down into steps, and decides on the next subtask.
As an action, the VLM outputs the trajectory parameters $a_i$ for the physical robot to execute.
DMPs then map these parameters into a controlled robot motion: ensuring the robot will reach the given goal and providing a continuous reference trajectory.
We hypothesize that this combination will be effective because DMPs capture a diverse set of complex motions through a small number of parameters, and that these parameters are semantically meaningful (e.g., a given weight might correspond to increased motion in the $x$ axis).

In this section we present our LMP formalism.
Our paradigm starts by generating a text summary of the state from image and depth observations (Section~\ref{sec:method_a}). 
Once the VLM is given this language information, it then \textit{decomposes} the overall task into the next subtask the robot should perform (Section~\ref{sec:method_b}).
Finally, the VLM \textit{generates} the goals and weights for low-level DMPs that will achieve this subtask, and the robot executes the DMPs in its physical environment (Section~\ref{sec:method_c}).
Across our experiments this pipeline was sufficient for completing most tasks on the first attempt.
However, if the robot fails, we also incorporate a refinement step where the user can give natural language corrections, and the robot uses those suggestions to improve its next attempt (Section~\ref{sec:method_judge}).
Our overall method is summarized in \fig{front}, \fig{method}, and Algorithm~\ref{alg:1}.

\begin{algorithm}[t]
\caption{Language Movement Primitives (LMPs)}\label{alg:1}
\begin{algorithmic}[1]
\Require task prompt $\tau$ and DMP grounding prompt $s_\mathcal{G}$
\State $\Pi \gets \varnothing$ \Comment{Set of completed subtasks}
\State $c \gets 0$ \Comment{Retry counter for each subtask}
\Repeat{}
    \State $o \gets \text{environment\_state}$
    \State $\theta \gets \pi_\text{class}(\circ \mid o)$ \Comment{Extract object poses and labels}
    \State $\varphi_{i} \gets
    \pi_\mathcal{D}\left(\circ \mid \tau, o, \theta, \Pi\right)$ \Comment{Decompose for next subtask}
    \State $(W_i, \Delta z, \Delta \psi) \gets \pi_\mathcal{G}\left(\circ \mid \varphi_{i}, s_\mathcal{G}, o, \theta\right)$ \\ \Comment{Generate DMP weights and goal offsets}
    \State $\text{DMP}(W_i, \Delta z, \Delta \psi)$ \Comment{Execute DMP in environment}
    \State $r \gets \text{judge}()$ \Comment{Collect any feedback}
    \If{$r \ne \varnothing$ and $c < 3$}
        \State update\_prompts$(r)$ \Comment{Add $r$ to existing prompts}
        \State $\text{reset}(\varphi_i)$ \Comment{Reset to start of subtask $i$}
        \State $c \gets c + 1$
    \Else
        \State $\Pi \gets \Pi \cup \varphi_i$
        \State $c \gets 0$ \Comment{Reset retry counter}
    \EndIf{}
\Until{$\varphi_i = \text{done}$}
\end{algorithmic}
\end{algorithm}

\subsection{From Observations to State Descriptions}\label{sec:method_a}

The robot's observations $o$ include the robot's joint position and RGB-D images of the environment.
We directly provide the RGB environment image as an input to the VLM policy.
However, we also translate our entire observation $o$ into a templated natural language description of the state for the VLM to reason over.
For example, ``there is an \texttt{[object label]} located at \texttt{[this position]} and oriented with \texttt{[this orientation]}.''
We achieve this automated segmentation by using an open-vocabulary classifier to identify objects in the environment:
\begin{equation}
    \theta = [(l, p)_i]_{0}^N \sim \pi_\text{class}\left(\circ \mid o\right)
    \label{eq:classifier}
\end{equation}
Here $\theta$ is the set of all segmented object information.
This includes $l$, the textual object labels, and $p$, their 3D position and orientation in the robot's coordinate frame.
The policy $\pi_\text{class}$ converts the observations into segmented data including the object's label and global position in 3D space.
In our experiments we use Gemini-Robotics ER \cite{team2025gemini} and LangSAM \cite{ravi2024sam, liu2024grounding}, but other approaches for object detection and localization are also suitable \cite{khanam2024yolov11, minderer2022simple, gu2021open}.
Note that most off-the-shelf classifiers output pixel coordinates. Using the camera intrinsics, depth measurements, and the calibrated camera-to-robot extrinsics, we back-project pixels into 3D and transform them into the robot frame.
Once $\theta$ is obtained, we can then use the information in $\theta$ to automatically populate a text description of the environment.
Overall, the VLM inputs both the RGB image and the text description from $\theta$.

\subsection{From State Descriptions to Decomposed Subtasks} \label{sec:method_b}

Now that the VLM has the task and state, we move towards outputting DMP weights for robot motion.
We divide this overall policy $\pi$ into two parts: \textit{decomposition} $\pi_\mathcal{D}$ and \textit{generation} $\pi_\mathcal{G}$.
In the decomposition, the VLM outputs a language description of the next subtask $\tau_i$.
Each subtask is a step towards the completed behavior.
Consider our motivating example of cleaning the plate: this task $\tau$ breaks down into three subtasks including grasping the sponge $\tau_1$, carrying it to the plate $\tau_2$, and wiping the plate $\tau_3$.
We want each subtask to correspond to something executable by our DMPs.
We therefore provide a \textit{template} for the decomposition policy $\pi_\mathcal{D}$ to complete at each high-level timestep.
The subtasks within this template can be of two forms.
First is \texttt{ACTION(object)}, where the robot acts on a single object (e.g., ``grasp the sponge'').
Second is \texttt{ACTION(object) TO (object)}, which is used for actions with multiple objects (e.g., ``carry the sponge to the plate'').
These templates ensure that the decompositions are anchored in subtasks related to scene objects and that subtasks are only involving one primary object that is manipulated, ensuring the right difficulty for each DMP.

With this framework in mind, the decomposition policy $\pi_\mathcal{D}$ is a foundational vision and language model of the form:
\begin{equation}
    \varphi_{i} \sim 
    \pi_\mathcal{D}\left(\circ \mid \tau, o, \theta, [\varphi_k]_0^{i-1}\right)
    \label{eq:decomposer}
\end{equation}
Here $[\varphi_k]_{0}^{i-1}$ represents the sequence of previously proposed subtask templates. 
At the start of the interaction this sequence is empty, and the sequence iteratively grows each time the decomposition policy is queried.
Looking at \eq{decomposer}, $\tau$ is the high level task, and $o$ and $\theta$ are the RGB image and natural language state description from Section~\ref{sec:method_a}.
The completed subtask template $\varphi_i$ output by this model describes the motion that the next DMP should complete.

\begin{figure*}[t]
    \centering
    \includegraphics[width=\linewidth]{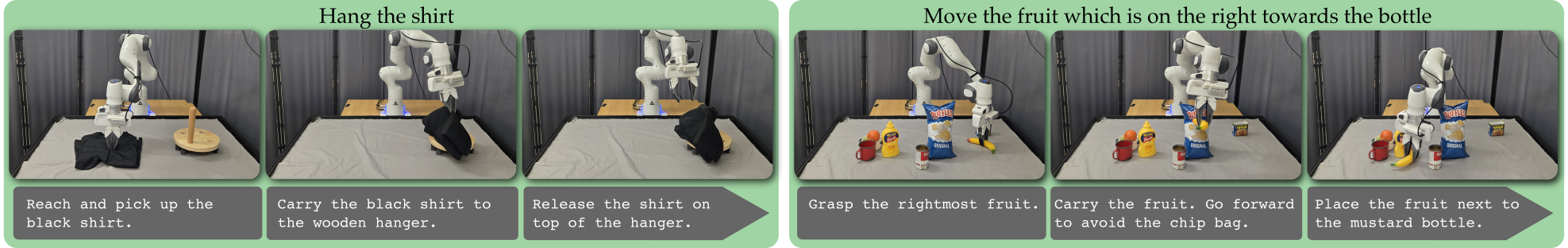}
    \caption{Our experiments evaluate LMP's performance on tabletop-manipulation tasks, converting natural-language task descriptions into robot controllers. In our tests, we evaluate 31 diverse household tasks requiring semantic task understanding, awareness of obstacles, and spatial reasoning. 
    }
    \label{fig:experiments}
\end{figure*}

\subsection{From Subtasks to Generating DMPs} \label{sec:method_c}

Now that we know the next desired motion, we ground this motion in the robot's physical environment with DMPs.
We specifically choose DMPs due to their semantically interpretable parameters: The forcing function $\mathbf{W})_i$ directly shapes the resulting trajectory, allowing for a deterministic change in the robot's motion along the modeled end-effector path. 
This property renders DMP weights as a natural action space for a VLM, allowing it to reason over and generate these weights from free-form task descriptions without requiring demonstrations or in-domain fine-tuning. 
Leveraging this semantic interpretability to ground VLM reasoning in continuous low-level motions is the central contribution of LMP, bridging symbolic task decomposition and embodied trajectory generation.

More specifically, given our scene description and next subtask, we sample from the \textit{generator policy} $\pi_\mathcal{G}$ to produce DMP weights $W_i$ and offset parameters $\Delta z$ and $\Delta \psi$:
\begin{equation}
    (W_i, \Delta z, \Delta \psi) \sim \pi_\mathcal{G}\left(\circ \mid \varphi_{i}, s_\mathcal{G}, o, \theta\right)
    \label{eq:generator}
\end{equation}
The generator VLM produces weights $W_i \in \mathbb{R}^{M \times B}$ where $M$ is the number of controlled degrees of freedom and $B$ denotes the number of basis functions used in the DMP formulation.
As a reminder, the basis functions are applied in \eq{P3} to shape the DMP trajectory. 
The details of the grounding prompt $s_\mathcal{G}$ will be outlined later in this subsection.

To improve tractability and reduce unnecessary complexity, we restrict the generator to the minimal set of operational dimensions required for task execution. 
Our experiments focus on top-down robot manipulation tasks where the end-effector is positioned above the target object.
We accordingly consider the motion of the robot end-effector in Cartesian space. 
Under this formulation, the weight matrix $W_i$ is composed of the following weight vectors:
\[
\mathbf{w_i}^{(x)},\ \mathbf{w_i}^{(y)},\ \mathbf{w_i}^{(z)},\ \mathbf{w_i}^{(\theta_z)},\ \mathbf{w_i}^{(g)} \in \mathbb{R}^{B}
\]
where $\mathbf{w_i}^{(x)}, \mathbf{w_i}^{(y)}, \mathbf{w_i}^{(z)}$ correspond to translational motion along the Cartesian axes, $\mathbf{w_i}^{(\theta_z)}$ represents the end-effector rotation about the $z$-axis, and $\mathbf{w_i}^{(g)}$ controls the gripper state. 
Note that the gripper command is represented as a continuous variable rather than a binary signal, allowing it to be controlled in the same manner as the other degrees of freedom.
In practice, we often just need the gripper to open or close at a specific timestep along the trajectory.
We encode this knowledge in the basis functions of the gripper DMP by replacing the Gaussian in \eq{P4} with a step function.
More generally, designers can modify the DMP hyperparameters (e.g., types and number of basis functions) to capture their domain knowledge.

The weights $W_i$ determine the continuous trajectory \textit{shape}.
But we also need to set the \textit{goal} $g$ that the DMP converges towards.
This goal is specified in two parts by offsets $\Delta z$ and $\Delta \psi$.
We first extract the location of the center of the object from $\theta$ in Section~\ref{sec:method_a}.
We then move towards a point aligned with the object's $xy$ position, but offset in both height and orientation.
The generator policy in \eq{generator} outputs a target height along the $z$-axis ($\Delta z$) and a target orientation around the $z$-axis ($\Delta \psi$).

Overall, the generator $\pi_\mathcal{G}$ provides the parameters that define DMPs in each Cartesian axis.
Our experiments show that we can use off-the-shelf vision and language models as $\pi_\mathcal{G}$, provided that we give these models some context on how DMPs operate.
We input this grounding through the task-invariant prompt $s_\mathcal{G}$.
This prompt (available in the supplemental material) encourages the generator to internalize the relationship between the physical workspace and the DMP weight space, i.e., how variations in DMP weights induce corresponding changes in Cartesian end-effector motion.
For example, the prompt explains that ``increasing the weight of parameters in the $x$-axis will cause the trajectory to move in that direction.''
Equipped with $s_\mathcal{G}$, the generator policy is able to translate subtask descriptions into low-level DMP parameters; hence, the subtask $\varphi_i$ only needs to be in natural, human-understandable language.

After decomposition and generation, we execute the DMP with weights $W_i$ and goal pose $\mathbf{g}$:
\begin{equation}
\mathbf{g} = 
\begin{bmatrix}
p_k^{x} & 
p_k^{y} & 
p_k^{z} + \Delta z & 
p_k^{\theta} &
p_k^{\phi} &
p_k^{\psi} + \Delta \psi
\end{bmatrix}
\end{equation}
Here $p_k$ is the position of the target object.
The position of all objects is originally estimated in $\theta$ from Section~\ref{sec:method_a}, and the target object is assigned by subtask $\varphi_i$.
The robot rolls out this motion in the environment.
More precisely, the robot uses the DMP as a reference trajectory, and outputs low-level control commands to track the reference.

\subsection{Using Feedback to Refine the Motion}\label{sec:method_judge}

Ideally, the motion produced in Section~\ref{sec:method_c} is correct and the robot completes the desired subtask.
But in practice this may not be the case; we therefore enable iterative, step-by-step feedback through a \textit{subtask judge}.
The subtask judge observes how the environment changes over the course of subtask $\varphi_i$, and provides qualitative feedback in text form to the decomposer and the generator.
The judge outputs a qualitative feedback statement $r$ that describes the error and/or how the robot should improve (e.g., ``too high, you missed the sponge'').
To incorporate this feedback and refine the robot's motion, we add the refinement $r$ to the robot's existing prompts for both the decomposer and the generator.
The environment is then reset to the configuration it was in \textit{before} the DMP was executed, and the decomposer and generator reason about appropriate corrections.
Through this in-context learning mechanism, we are able to iteratively improve both the task decomposition and generated weights, leading to safer and more efficient interactions.
\section{Experiments}

\begingroup
\renewcommand{\arraystretch}{1.0} 
\begin{table*}[]
    \centering
    \small
    \setlength{\tabcolsep}{0.5em}%
    \caption{Tasks and Performances.\\ Reported in success rate ($\%$). Numbers in parenthesis $(\circ)$ represent the number of feedbacks per task, averaged across trials. \\
    Columns $4$-$6$ refer to our judge-free (J/F), decomposer free (D/F), and judge- \& decomposer-free (J/F D/F) ablations.
    \\ 
    Task Categories:
    \ \textcolor{myred}{\Large$\bullet$} Semantic understanding~ \textcolor{mygreen}{\Large$\bullet$} Obstacle awareness~ 
    \textcolor{myblue}{\Large$\bullet$} Spatial reasoning}
    \label{tab:results}
    \begin{tabular}{ r|c|c|c|cccccc}
    \toprule
         \multicolumn{1}{r|}{ \raisebox{0.3em}{\textbf{Task Name}}} & \textcolor{myred}{\Huge$\bullet$} & \textcolor{mygreen}{\Huge$\bullet$} & \textcolor{myblue}{\Huge$\bullet$} & \raisebox{0.3em}{\textbf{LMP}} & \shortstack{\textbf{TrajGen} \\ \cite{Kwon2024Language} } & \shortstack{\raisebox{0.3em}{$\pi_{0.5}$} \\ \cite{intelligence2025pi_} } & \shortstack{\textbf{LMP} \\J/F}
& \shortstack{\textbf{LMP} \\ D/F} & \shortstack{\textbf{LMP} \\ J/F D/F}\\ \midrule
    pick the chip bag on 
    the left of the table 
    & & & \ding{51} & $\mathbf{100}~(0.2)$ & $40$ & $20$ & $80$ & $40~(2.2)$& $20$
    \\
    pick the rightmost can 
    & & & \ding{51} & $\mathbf{100}~(-)$ & $20$ & $40$ & $\mathbf{100}$ & $20~(2.4)$& $20$ \\
    pick the fruit in the middle 
    & & & \ding{51} & ${60}~(1.2)$ & ${60}$ & $40$ & ${60}$ & $\mathbf{80}~(0.6)$ & $60$\\
    pick the chip bag which 
    is to the right of the can 
    & & & \ding{51} & $\mathbf{100}~(-)$ & $60$ & $20$ & $\mathbf{100}$ & $60~(0.6)$ & $60$\\
    move the fruit which is on
     the right towards the bottle 
    & & & \ding{51} & $\mathbf{100}~(-)$ & $60$ & $0$ & $\mathbf{100}$ & $80~(0.6)$ & $60$\\ 
    \rowcolor[gray]{0.925} move the banana near the pear 
    & & & \ding{51} & $\mathbf{100}~(-)$ & $\mathbf{100}$ & $60$ & $\mathbf{100}$ & $80~(0.6)$ & $40$\\
    \rowcolor[gray]{0.925} move the banana near the pear 
    (obstacles included) 
    & & \ding{51} & \ding{51} & $\mathbf{80}~(1.0)$ & $20$ & $60$ & $40$ & $0~(3)$ & $0$\\
    \rowcolor[gray]{0.925} move the can to the 
    center of the table 
    & & & \ding{51} & $\mathbf{80}~(0.6)$ & $20$ & $20$ & $\mathbf{80}$ & $\mathbf{80}~(0.6)$& $40$\\
    \rowcolor[gray]{0.925} move the lonely object
    to the others
    & \ding{51} & & \ding{51} & ${80}~(0.6)$ & $0$ & $60$ & ${80}$ & $\mathbf{100}~(0.2)$& $80$\\
    \rowcolor[gray]{0.925} place the apple in the bowl 
    & & & & $\mathbf{60}$ (1.2)& $\mathbf{60}$ & $40$ & $\mathbf{60}$ & ${40}~(0.6)$ & $0$\\ 
    place the apple in the bowl (obstacles included) 
    & & \ding{51} & & $\mathbf{60}~(1.6)$ & $0$ & $20$ & 20 &  $\mathbf{60}~(1)$& $0$\\
    pick the apple from the bowl 
    and place it on the table 
    & & & & $\mathbf{80}~(0.6)$ & $60$ & $40$ & $\mathbf{80}$ & $20~(1.2)$ & $20$\\
    wipe the plate 
    & \ding{51} & & & $\mathbf{80}~(0.6)$ & $0$ & $0$ & $\mathbf{80}$ & $0~(3)$ & $0$\\
    drop the ball into the cup 
    & & & & $\mathbf{60}~(1.4)$ & $0$ & $20$ & $40$ & $20~(1.2)$ & $0$\\
    drop the ball into the cup (obstacles included) 
    & & \ding{51} & & $\mathbf{80}~(2.4)$ & $0$ & $0$ & $20$ & $20~(1.2)$ & $0$\\ 
    \rowcolor[gray]{0.925} insert the bread into the toaster 
    & & & & $\mathbf{80}~(0.6)$ & $0$ & $0$ & $\mathbf{80}$ & $20~(2.2)$ & $0$\\
    \rowcolor[gray]{0.925} pick up the bowl 
    & & & & ${80}~(0.6)$ & $20$ & $40$ & ${80}$ & $\mathbf{100}~(0.4)$ & $60$\\
    \rowcolor[gray]{0.925} hang the shirt
    & \ding{51} & & & $\mathbf{40}~(1.8)$  & $0$ & $\mathbf{40}$ & $\mathbf{40}$ & $\mathbf{40}~(0.6)$ & $20$\\
    \rowcolor[gray]{0.925} put the banana on the plate 
    & & & \ding{51} & $\mathbf{80}~(0.6)$ & $\mathbf{80}$ & $60$ & $\mathbf{80}$ & $60~(1.2)$ & $60$\\
    \rowcolor[gray]{0.925} put the banana on the plate (obstacles included) 
    & & \ding{51} & \ding{51} & $\mathbf{100}~(0.4)$ & $0$ & $40$ & $60$ & $40~(3)$ & $0$\\
    knock over the left bottle
    & & & \ding{51}& $\mathbf{100}~(0.2)$& ${80}$& ${60}$& ${80}$& ${80}~(1.6)$& ${60}$\\
    push the bottle to the orange
    & & & \ding{51}& $\mathbf{40}~(1.2)$& $\mathbf{40}$& $\mathbf{40}$& $\mathbf{40}$& $\mathbf{40}~(2.2)$& ${0}$\\
    push the can towards the right
    & & & \ding{51}& $\mathbf{60}~(1.2)$& ${20}$& ${0}$& $\mathbf{60}$& $\mathbf{60}~(1.2)$& $\mathbf{60}$\\
    stir the mug with the spoon
    & \ding{51}& & & $\mathbf{40}~(1.6)$& ${0}$& ${0}$& ${0}$& ${20}~(2.6)$& ${0}$\\
    draw a five-pointed star
    & & & \ding{51}& ${0}~(3)$& $\mathbf{100}$& ${0}$& ${0}$& ${0}~(3)$& ${0}$\\
    \rowcolor[gray]{0.925} move the pan to the left
    & & & \ding{51}& $\mathbf{60}~(1.4)$& ${40}$& ${0}$& ${40}$& ${40}~(1.8)$& ${40}$\\
    \rowcolor[gray]{0.925} wipe the table with the sponge 
    & & & \ding{51}& $\mathbf{80}~(0.8)$& ${60}$& ${60}$& ${60}$& ${60}~(1.8)$& ${40}$\\
    \rowcolor[gray]{0.925} draw a circle with the gripper closed
    & & & \ding{51}& ${0}~(3)$& $\mathbf{100}$& ${40}$& ${0}$& ${0}~(3)$& ${0}$\\
    \rowcolor[gray]{0.925} unplug the charger
    & \ding{51}& & & ${20}~(2.4)$& ${0}$& ${20}$& ${20}$& $\mathbf{40}~(2.8)$& ${0}$\\
    \rowcolor[gray]{0.925} take out the tissue from the dispenser
    & \ding{51}& & & ${0}~(3)$& $\mathbf{20}$& $\mathbf{20}$& ${0}$& ${0}~(3)$& ${0}$\\
    open the bottle cap
    & \ding{51}& & & $\mathbf{20}~(3)$& $\mathbf{20}$& ${0}$& ${0}$& ${0}~(3)$& ${0}$
    \\
\bottomrule
\multicolumn{4}{r|}{\textbf{Overall Performance}} 
& \textbf{65} & $35$& $28$& $54$& $41$& $24$\\
\bottomrule
     \end{tabular}
\end{table*}
\endgroup 

To demonstrate the effectiveness of LMP, we compare it to state-of-the-art baselines and conduct extensive ablations in a controlled tabletop manipulation setting involving household objects. 
We evaluate LMP on $31$ household manipulation tasks (\fig{experiments}), covering the full set of tasks evaluated in~\cite{Kwon2024Language}. We compare overall task success rates against $\pi_{0.5}$~\cite{intelligence2025pi_} and TrajGen~\cite{Kwon2024Language}.
Furthermore, we ablate our method to evaluate the impact of free-form natural language feedback on motion generation as well as the impact of grounding subtask identification through task decomposition.

\p{Setup}
We evaluate these methods on a 7-DoF Franka Emika Panda robot arm with a UMI gripper (see \fig{experiments}). We use a mounted Orbbec Femto Mega camera for RGB-D imagery. 
For $\pi_{0.5}$, we instead use two Intel Realsense D435 cameras: one mounted on the robot end-effector and one statically mounted with an isometric view of the workspace, covering the same field of view. 
We use Gemini Robotics-ER 1.5~\cite{team2025gemini} to obtain object labels $\{l_i\}_{i=1}^M$. 
These labels are then passed to LangSAM~\cite{liu2024grounding, ravi2024sam}, which produces fine-grained object segmentations.
These segmentation masks are subsequently projected onto the depth image to recover the corresponding 3D object bounding boxes. 
We use these bounding boxes to determine the approximate yaw of each object relative to the end-effector orientation.
Leveraging its strong capabilities in physical scene understanding and multimodal reasoning for robotic tasks, we also employ Gemini Robotics-ER as our task decomposer.
We use GPT-5.2~\cite{openai_gpt5_2_system_card_2025} as the generator policy. 
Feedback, when needed, is provided by a single person to ensure consistency across the evaluation. 

\p{Tasks} 
For our experiments, we choose a set of $31$ tasks, inspired by~\cite{Kwon2024Language}, which contain a diverse set manipulation challenges including spatial reasoning, semantic scene understanding, and obstacle awareness (see \fig{experiments}). 

In contrast to~\cite{Kwon2024Language}, we introduce task variants that require the robot to reason about object interactions and potential collisions along its motion, which are not explicitly mentioned in the task instructions, but are implicitly relevant.
To further evaluate our method’s ability to interpret semantically complex tasks, we provide generalized task descriptions. For example, we use instructions such as ``Wipe the plate'' instead of ``Wipe the plate with a sponge.'' In these cases, we rely on the decomposer to infer the implicit steps and complete the full sequence of actions.


We expect language-enabled motion generators to exhibit three core capabilities: (1) semantic understanding, inferring intent from under-specified instructions; (2) obstacle awareness, recognizing configurations the robot must avoid; and (3) spatial reasoning, understanding spatial relations between objects in the environment. We label tasks according to these characteristics to analyze method capabilities.



\p{Metrics} 
We employ two performance metrics across tasks:
\begin{enumerate}
    \item Success rate: For each task, the success rate is evaluated over 5 independent trials. In each trial, the positions and orientations of the target objects, as well as the obstacles and distractors, are randomly initialized. We only consider a task successful if all constituent subtasks are completed successfully and no collisions occur.
    \item We analyze the failure modes of the model with respect to five common categories.
\end{enumerate}

\subsection{Baseline Comparison}

We compare our approach against two representative state-of-the-art baselines that use pretrained language models to guide robot policies from natural language instructions. 
Our selection of baselines reflects two conventions by which language models are incorporated into robotics: as common sense reasoning agents and language-grounded action generators.  

\p{TrajGen~\cite{Kwon2024Language}}
\cite{Kwon2024Language} leverages the common sense reasoning capabilities of LLMs to generate executable motion representations. 
Specifically, TrajGen employs an LLM to synthesize code that generates dense waypoint trajectories based in the task instruction, which are subsequently executed by the robot.
To ensure a consistent evaluation, we provide TrajGen with the same information from the object detection pipeline and employ GPT-5.2 as its internal LLM.

\p{Pi-0.5~\cite{intelligence2025pi_}}
Vision-Language-Action models such as $\pi_{0.5}$ leverage pretrained language models to effectively ground language instructions in action and visual perception.
This allows VLAs to generalize across tasks by encoding high-level semantic knowledge acquired during large-scale pretraining ~\cite{zitkovich2023rt}. We train $\pi_{0.5}$ using an aggregated demonstration dataset across all $31$ tasks, with five demonstrations per task. 
We follow the officially released Libero fine-tuning scheme~\cite{intelligence2025pi_} and fine-tune from the standard checkpoint for $20000$ steps.

\p{Baseline Results} We report the performance of LMP when compared to the baselines in Table~\ref{tab:results}. 
We observe that LMP consistently outperforms TrajGen and $\pi_{0.5}$ (col. $1$-$3$) with an overall success rate of $65\%$ as compared to $35\%$ and $28\%$ for TrajGen and $\pi_{0.5}$ respectively.
We attribute this improvement to decoupling high-level semantic reasoning from low-level motion generation: the semantically interpretable parameter space allows general-purpose foundation models to specify task intent without requiring knowledge of robot dynamics, while the low-level controller handles trajectory execution.

We observe that TrajGen particularly struggles with tasks that require nonlinear motions, such as ``wipe the plate'', which requires the robot to move in a circular pattern. 
TrajGen tends to output linear motion, whereas LMP is capable of generating smooth splines. 
Additionally, in tasks that require semantic understanding, TrajGen often neglects task-relevant objects that are not directly specified in the task description. 

%
We also found that $\pi_{0.5}$ experiences causal confusion \cite{dai2025civil}. For example, if an orange is always near an apple during data collection, $\pi_{0.5}$ will fail to develop a strong semantic distinction between these two objects. 
We also observe that when the task-relevant entities do not provide a strong visual signal, such as referencing the ``middle of the table'' as opposed to ``the plate'' as a target position, it is more challenging for $\pi_{0.5}$ to complete the task.
This observation is common when fine-tuning a vision-language backbone for low-level control, as it diminishes its overall reasoning capability~\cite{zhou2025chatvla, dey2025revla}.

%
%
%
%

\begin{figure}
    \centering
    \includegraphics[width=1\linewidth]{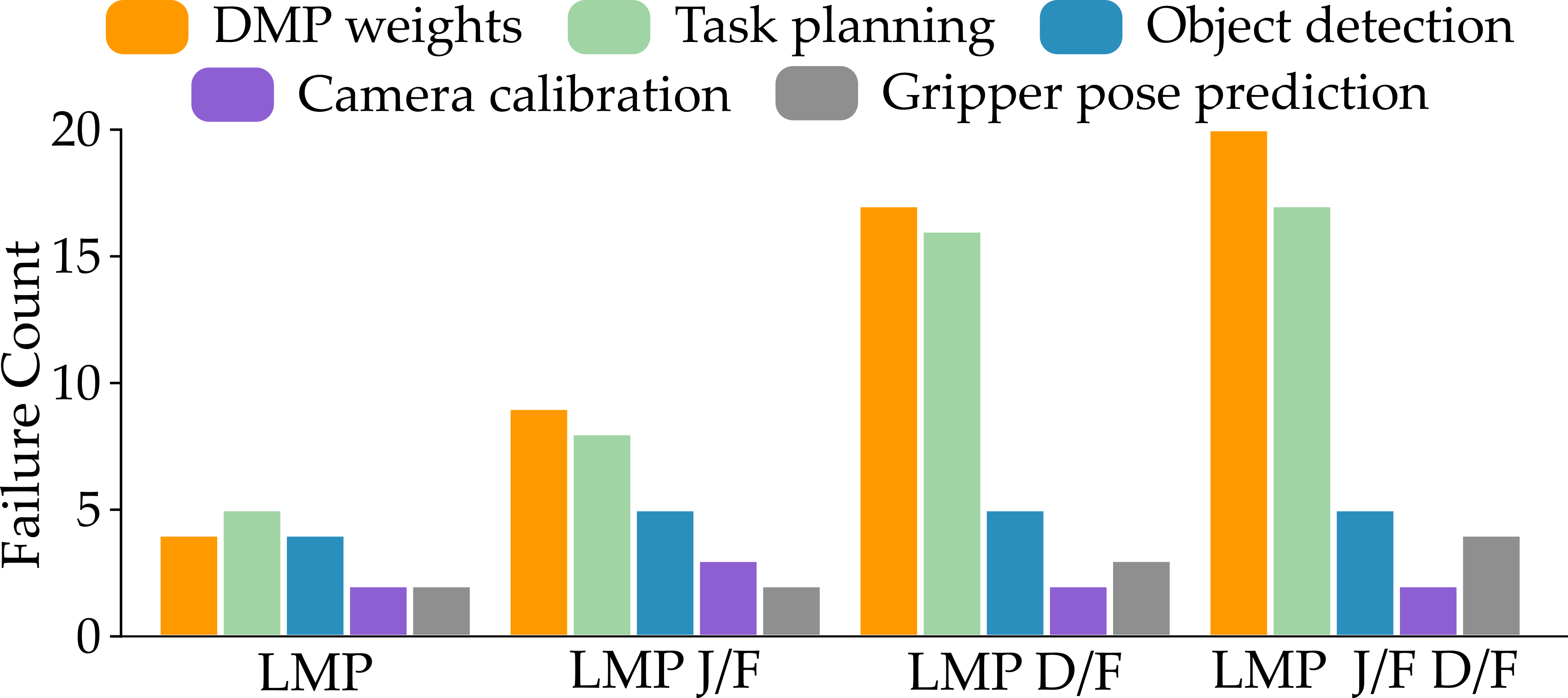}
    \caption{We identify five dominant failure modes and analyze the impact of the judge as well as the task decomposition on these failure modes across task executions. 
    Particularly the addition of a grounded task-decomposition substantially reduces the errors in task planning and subsequently more effective weight generation.
    Furthermore, we find that adding feedback has a strong influence on successful DMP weight generation. 
    }
    \label{fig:failure}
\end{figure}

\subsection{Ablation Study}

Two key components of our method are the ability to incorporate feedback and the utilization of a task decomposition module.
To assess the impact of these two components, we conducted extensive ablation studies (see Table~\ref{tab:results}, col $4$-$6$).

\p{Judge-Free Ablation (J/F)} We test our method without the judge feedback outlined in Section~\ref{sec:method_judge}. 
This ablation (LMP J/F) focuses the success of generating DMP weights on the first try, removing the ability to refine the motion through feedback.

\p{Decomposer-Free Ablation (D/F)}
We also evaluate our method without the task decomposer described in Section~\ref{sec:method_b} (LMP D/F). In this setting, the generator VLM is prompted to infer a sequence of DMPs, goals, and offsets, decomposing the task implicitly. This ablation aims to demonstrate the importance of a dedicated decomposer to anchor the DMP generation in appropriate subtask complexity.

\p{Judge- \& Decomposer-Free Ablation (J/F D/F)}
We evaluate a further ablation (LMP J/F D/F) of \textit{Decomposer-Free} by removing expert feedback from the generator policy. This ablation intends to show that expert feedback is especially critical when the generator lacks structured subtask grounding. 

\p{Ablation Results}
We find that removing either the judge or the decomposer leads to reduced performance. J/F achieves a success rate of $54\%$, while removing the decomposer D/F degrades performance to $41\%$.
Removing both results in a success rate of $24\%$.
The sources of failures differs across ablations, as shown in \fig{failure}.
We find that both the judge and decomposer are necessary components to improve \textit{DMP Weight Generation} and \textit{Task Planning}. 
Removing the decomposer leads to a major degradation in overall task performance.
Without it, LMP tends to manipulate task-irrelevant objects and reason about the task at an inappropriate level of abstraction: either too complex to be effectively captured by DMP parameterization, or too granular, at which point general-purpose foundation models lack the required knowledge to successfully reason over embodied robot dynamics. 
%
Furthermore, the judge particularly improves weight generation by allowing LMP to refine DMP parameters.



In summary, these ablations highlight the complementary roles of the judge and decomposer in grounding the foundation model at an appropriate level of abstraction.

\begin{figure*}[t]
    \centering
    \includegraphics[width=\linewidth]{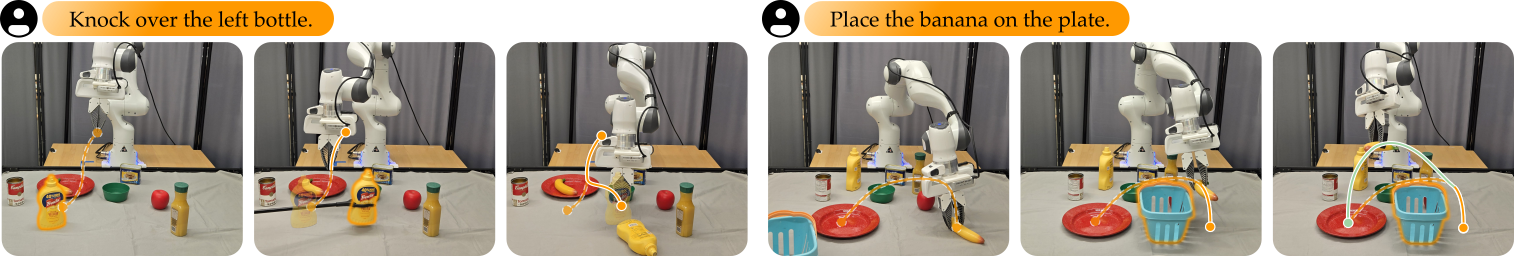}
    \caption{We evaluate LMP in dynamic environments where object positions and scene configurations may change during execution. \textit{Left:} The target object is moved during execution. In this case, LMP does not need to regenerate DMP weights, and instead adapts by updating the target goal through the goal-attractor dynamics of the DMP. \textit{Right:} New obstacles are introduced along the robot's path. In this scenario, LMP updates the DMP weights in closed loop to reshape the trajectory and avoid the obstacle.
    }
    \label{fig:dynamic}
\end{figure*}

\begin{figure}
    \centering
    \includegraphics[width=1\linewidth]{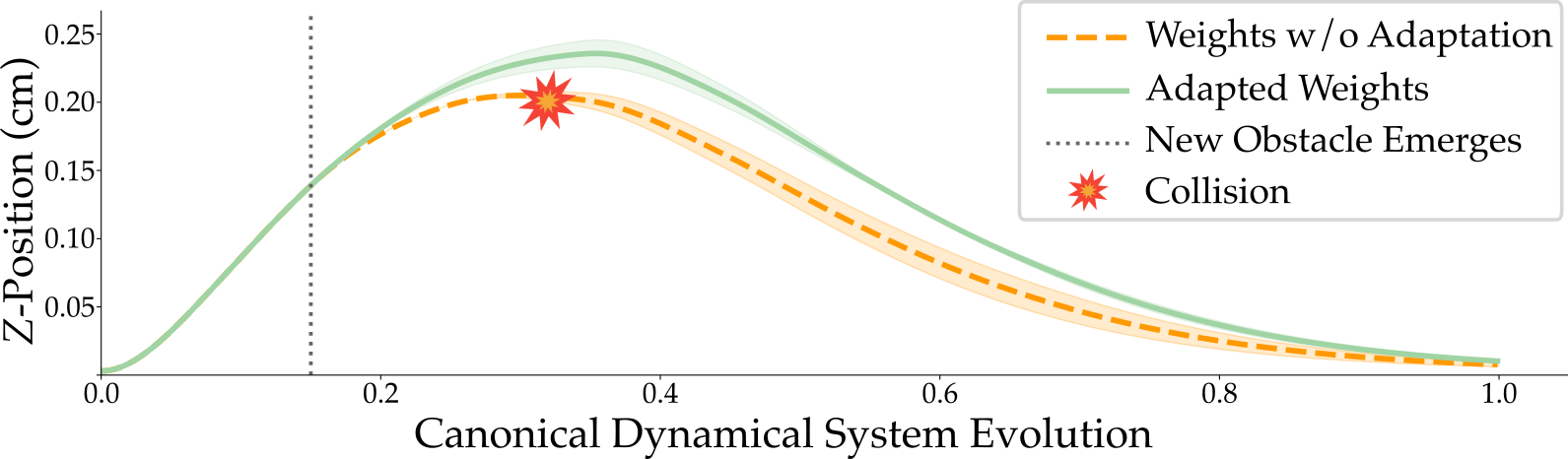}
    \caption{LMP with support for real-time weight adaptation is able to reactively adjust to changes in the environment during rollout, such as the introduction of new obstacles, to avoid collision while still converging to the goal position and completing the task.
    }
    \label{fig:adapt}
\end{figure}

\subsection{Reactive Adaptation}
Reactive adaptation to changing environments is crucial in task and motion planning. Object positions and scene configurations may evolve during execution, and robotic systems should be able to adapt to these changes in real-time. 
To this end, we evaluate a version of LMP with support for real-time adaptation (LMP-Adapt).
This version demonstrates that LMs are able to reason over the weights of DMPs when shifting targets and changes in obstacle configurations occur. Specifically, we compare LMP-Adapt against baseline methods on two categories of dynamic tasks (Table~\ref{tab:dynamic}).

\p{Moving target objects} In these tasks, we evaluate scenarios with dynamically changing goal positions $g$ while preserving the overall motion structure generated by the VLM. We use a video segmentation and tracking model~\cite{cheng2023tracking} to track the objects involved in the current subtask and update the DMP goal position online during execution. This setting is designed to demonstrate that the goal-attractor dynamics of DMPs lends itself well to dynamic environments.

\p{Dynamic obstacles} In these tasks, a VLM continuously incorporates new observations during rollouts and updates the DMP weights in a closed loop manner, enabling responsive replanning as the environment evolves.
Importantly, while the goal position remains the same, this task evaluates whether or not the shape of the motion itself adapts to obstacles that need to be avoided during the execution of the motion. 
DMPs lend themselves particularly well to this setting as their goal-attractor dynamics continue to converge to the goal, while the weights of the forcing terms allow the VLM to re-shape the motion itself in the semantically interpretable parameter space to avoid the obstacle.
We consider two scenarios that alter the trajectory shape: the introduction of a new obstacle along the robot's path, and the removal of an existing obstacle during execution. Specifically, we use GPT-5.4-nano~\cite{openai2026gpt54} as a lightweight VLM that provides interpretable outputs while supporting closed-loop updates at a reasonable frequency of approximately $0.3$~Hz. This setting is designed to demonstrate that, despite being bottlenecked by language model inference time, LMP with weight adaptation is still capable of performing closed-loop control and producing semantically meaningful and reasonable trajectory adjustments in response to environmental changes.

Our results in Table~\ref{tab:dynamic} and Table~\ref{tab:dynamic_obstacles} show that LMP outperforms the baselines across both categories of dynamic tasks. In the moving target object scenarios, TrajGen is unable to adjust its trajectory because its execution is fully open loop. In contrast, while $\pi_{0.5}$ demonstrates certain ability to track moving objects, we observe that it is more prone to going out of distribution under visual disturbances and changing scene configurations, causing an overall performance drop of $20\%$. 
In the dynamic obstacle scenarios, we observe that LMP-Adapt is sometimes unable to adjust its trajectory quickly enough in response to environmental changes due to the limited update frequency of the VLM (recall that we are using $0.3$~Hz for the live updates).
Nevertheless, across the two tasks we evaluate, LMP-Adapt does not show a significant drop in success rate compared to the static scene setting, retaining a performance of $55\%$, whereas $\pi_{0.5}$ show noticeable performance degradation to $5\%$. We show LMP bending the trajectory shape along the $z$-axis in response to emerging obstacles across rollouts in Figure~\ref{fig:adapt}. 

From the per-DoF changes presented in Table~\ref{tab:dynamic_obstacles}, we observe that the weight modifications induced by the language model in response to dynamic objects primarily affect the $z$ direction, which is expected as the robot mainly moves in the $z$-axis to avoid objects. We also observe changes along the $y$-axis.
However, in contrast to the $z$-axis, the motion along the $y$ axis mainly influences the temporal progression of the motion to allow for the spatial adjustment of the $z$-axis. 
We test this hypothesis by applying Dynamic Time Warping (DTW)~\cite{1163055}  in addition to plainly computing the distance between corresponding trajectory points.  
We observe that the accumulated deviation along $y$ over 1000 timesteps remains relatively small ($9.238 \pm 7.700 \,\mathrm{cm}$) compared to the much larger changes observed along $z$ ($92.06 \pm 156.8 \,\mathrm{cm}$), supporting our observation that changes in $y$ are largely temporal, not spatial adjustments. 
In particular, when obstacle configurations change, the adapted $y$-axis weights enable the robot to slow down and temporally coordinate its motion with the concurrent height adjustments along the $z$-axis required for obstacle avoidance. 
These results suggest that the language model is capable of reasoning over spatial relationships and expressing such reasoning through meaningful adaptations of the DMP trajectory weights.

\begingroup
\renewcommand{\arraystretch}{1.0}

\begin{table}[]
    \centering
    \scriptsize
    \setlength{\tabcolsep}{0pt}
    \caption{Reactive Planning Performances. Reported in success rate ($\%$). \textbf{Bolded objects} are being moved during the rollouts.}
    \label{tab:dynamic}

    \begin{tabular*}{\columnwidth}{@{}r@{\hspace{0.6em}}|@{\hspace{2.2em}}c@{\hspace{4.5em}}c@{\hspace{4.5em}}c@{}}
    \toprule
    \textbf{Moving Target Objects}
    & \textbf{LMP}
    & \textbf{TrajGen} \cite{Kwon2024Language}
    & $\pi_{0.5}$ \cite{intelligence2025pi_}\\
    \midrule
    place the apple in the \textbf{bowl} & 60 & 0 & 20 \\
    wipe the \textbf{plate} & 60 & 0 & 20 \\
    hang the shirt \textbf{(hanger)} & 20 & 0 & 0 \\
    knock over the \textbf{left bottle} & 80 & 0 & 0 \\
    stir the \textbf{mug} with the spoon & 20 & 0 & 0 \\
    \midrule
    \textbf{Overall Performance} & \textbf{48} & 0 & 8 \\
    \bottomrule
    \end{tabular*}

    \vspace{-0.5em}
\end{table}

\begin{table}[]
    \centering
    \scriptsize

    \caption{Dynamic obstacle performance and per-DoF changes. Performance is reported as success rate ($\%$); Per-DoF changes (cm) over 1000 timesteps, computed as the accumulated absolute timestep-wise difference between trajectories generated from the original and adapted weights.}
    \label{tab:dynamic_obstacles}

    \begin{tabular*}{\columnwidth}{@{\extracolsep{\fill}}r@{\hspace{0.6em}}|@{\hspace{0.1em}}c@{\hspace{0.2em}}c@{\hspace{0.2em}}c@{\hspace{0.2em}}c@{}}
    \toprule
    \raisebox{0.45em}{\textbf{Dynamic Obstacles}}
    & \shortstack{\textbf{LMP-}\\\textbf{Adapt}}
    & \raisebox{0.45em}{\textbf{LMP}} 
    & \shortstack{\textbf{TrajGen}\\ \cite{Kwon2024Language}} 
    & \shortstack{$\pi_{0.5}$\\ \cite{intelligence2025pi_}} \\
    \midrule
    \textbf{T1.} put the banana on the plate \textbf{(emerging obstacle)} & 40 & 0 & 40 & 20 \\
    \textbf{T2.} put the banana on the plate \textbf{(obstacle removed)} & 100 & 80 & 80 & 0 \\
    \textbf{T3.} drop the ball into the cup \textbf{(emerging obstacle)} & 20 & 0 & 0 & 0 \\
    \textbf{T4.} drop the ball into the cup \textbf{(obstacle removed)} & 60 & 60 & 0 & 0 \\
    \midrule
    \textbf{Overall Performance} & \textbf{55} & 35 & 30 & 5 \\
    \bottomrule
    \end{tabular*}

    \vspace{0.25em}

    {
    \setlength{\tabcolsep}{1pt}

    \begin{tabular*}{\columnwidth}{@{\extracolsep{\fill}}c@{\hspace{0.05em}}|cccc@{}}
    \toprule
    \textbf{} & x & y & z & yaw \\
    \midrule
    \textbf{T1.} 
    & $117.7{\pm}235.5$ 
    & $409.5{\pm}358.1$ 
    & $\mathbf{790.9{\pm}587.7}$ 
    & $0.000{\pm}0.000$ \\

    \textbf{T2.} 
    & $94.03{\pm}188.1$ 
    & $473.6{\pm}116.1$ 
    & $\mathbf{1092{\pm}22.54}$ 
    & $0.000{\pm}0.000$ \\

    \textbf{T3.} 
    & $40.97{\pm}2.284$ & $551.3{\pm}25.80$ & $\mathbf{1098{\pm}9.824}$ & $1.236{\pm}0.07068$ \\

    \textbf{T4.} 
    & $68.39{\pm}136.8$ 
    & $615.1{\pm}2.630$ 
    & $\mathbf{1080{\pm}0.001}$ 
    & $4.202{\pm}7.352$ \\
    \bottomrule
    \end{tabular*}
    }

    \vspace{-0.5em}
\end{table}

\endgroup

\subsection{Long-Horizon Benchmark}

To show that LMP can effectively handle long-horizon manipulation tasks, we evaluate it on a subset of the CALVIN tasks~\cite{mees2022calvin}.
Specifically, compared to the original CLAVIN task, we use a modified version that consisting of a preliminary subset of five representative manipulation tasks from the benchmark, including pick-and-place, pushing, and pulling behaviors.
LMP achieves promising sequential success rates across five randomly ordered CALVIN tasks (Table~\ref{tab:calvin_results}), demonstrating easy adaptation to novel benchmarks beyond pick-and-place tasks while using the same task-invariant grounding prompt. 
Particularly for the long-horizon case with a sequence of five subtasks, LMP is capable of chaining and generating motions from the result of prior executions, resulting in an overall zero-shot success rate of $42.8\%$.
These results further highlight the scalability of LMP, showing that the framework can generalize to more diverse manipulation settings without task-specific redesign.

\begin{table}[]
    \centering
    \caption{CALVIN experiments; 5 task chains \\ Tasks: open drawer; slide door left/right; place cube in drawer, close drawer, turn on/off led}
    \begin{tabular}{r|cc}
         \midrule
         \multicolumn{3}{c}{$\varnothing \to$ D  $\vert$ S1: $100\%$ $\vert$ S2: $85.7\%$ $\vert$ S3: $71.4\%$ $\vert$ S4: $42.8\%$ $\vert$ S5: $42.8\%$} \\
    \end{tabular}
    \label{tab:calvin_results}
    \vspace{-2.0em}
\end{table}

\subsection{Other Action Representations}

We use DMPs to model the end effector's position and orientation in Euler-angle space throughout our evaluations. However, the core insight of VLM-based controller generation is not limited to DMPs. 
Fundamentally, the key insight in LMP is that semantically interpretable parameters provide a capable interface to connect LLM and VLM based reasoning with embodied low-level control. 
To this end, the following demonstrates adaptations to different action spaces as well as a discussion of alternative controllers.

In terms of action parameterization, besides using Euler angles for rotation representation, we also explore quaternions as an alternative expressive rotational parameterization. Across $11$ of the \textit{spatial reasoning} tasks, chosen as a subset from Table~\ref{tab:results} for their need for rotational alignment, LMP maintains strong performance under both representations, achieving overall success rates of $82\%$ and $89\%$ with quaternion and Euler-angle parameterizations respectively. These results demonstrate the adaptability of LMP to different action representations.

In terms of controller formulation, LMP currently focuses on point-attractor DMPs, while extensions incorporating alternative controllers are left for future work. Other trajectory generation methods, such as B-spline parameterizations~\cite{zhou2026beast}, could also be integrated within the same framework. We choose DMPs over alternative trajectory representations for several reasons. A key motivation is that many manipulation tasks, such as top-down grasping, wiping, and obstacle avoidance, fundamentally require shaped motions, where directly moving toward a goal attractor would be infeasible. DMPs naturally support such structured trajectories through their learned forcing functions, while still maintaining stable goal-directed behavior. In addition, DMP weights carry temporally grounded semantics, making them particularly amenable to reasoning and generation by VLMs. 

Another important advantage is that the goal-attractor dynamics of DMPs enable online adaptation to moving targets through simple updates to the goal state $g$, retaining the overall motion structure generated by the VLM. Such reactive adaptation is less straightforward to achieve with geometric trajectory representations such as B-splines. In practice, the generated DMP is executed by sampling the trajectory at a fixed frequency and leveraging the robot's position-control API. Furthermore, the closed-loop nature of DMPs~\cite{schaal2006dynamic, li2023prodmp} provides temporal corrections through error calculation and stable convergence behavior, as the influence of the forcing function gradually decays to zero over the course of execution.

\section{Conclusion} \label{sec:conclusion}

We presented a framework to bridge the gap between language understanding and embodied motion control.
Our core contribution is a novel framework to ground the common-sense reasoning capabilities of vision-language models into embodied motion through control-theoretic approaches leveraging dynamic movement primitives.
We argue that this combination is effective because DMPs have a relatively small number of semantically meaningful parameters that VLMs can inherently tune.
Through extensive experiments in static and dynamic environments, we demonstrate that the VLM is cable to specify the necessary weights, thereby translating high-level reasoning into low-level embodied behaviors. 
We evaluate our approach through $31$ tabletop manipulation tasks:
when using LMP, the robot completed tasks on its first attempt $54\%$ of the time, and $65\%$ of the time when given up to $3$ rounds of natural language corrections.
This contrasts with VLM approaches that do not incorporate DMPs ($35\%$ success rate), and with robotics foundation models that need in-domain fine-tuning ($28\%$ success rate).

\p{Limitations and Future Work}
While LMP provides a framework that bridges the gap between high-level semanic reasoning and low-level control, it requires the low-level controller to have a set of semantically intepretable parameters. 
While this work focuses on the use of DMPs, we hypothesize that the proposed framework can be extended in future work to incorporate other classes of controllers as long as their respective parameters are semantically interpretable for a VLM. 
Furthermore, live updates in dynamic environments are limited by how fast a VLM can respond (currently $\approx 0.3$Hz).
Future work needs to be done to assess the framework’s ability to reason over relative spatial relationships. In particular, we observe that the language model fails more frequently when grounding trajectory weights from abstract relative instructions such as ``$10\,\mathrm{cm}$ to the left''.
Moreover, our evaluation is currently limited to five degrees of freedom. Extending the framework to higher-dimensional action spaces would introduce additional challenges for language models to reason over the motion, but represents an important direction for future investigation with will also scale with further advances in VLMs themselves.
Finally, while feedback from a human judge has proven useful, we will investigate the use of an autonomous judge, such as a VLM, in future work. 

%
%



\bibliographystyle{IEEEtran}
\bibliography{references}



\end{document}